\documentclass[letterpaper, 10 pt, conference]{ieeeconf}  % Comment this line out if you need a4paper

\IEEEoverridecommandlockouts                              % This command is only needed if 
\usepackage{amsmath}
\usepackage{booktabs}
\usepackage{graphicx}   % for including images
\usepackage{subcaption} % for subfigures (recommended modern way)
\usepackage{cite}
\usepackage{amsfonts}
\usepackage{amssymb}
\usepackage{algorithmic}
\usepackage{algorithm}
\usepackage{bm}

\usepackage{xcolor}
\usepackage{soul}
\newcommand{\remove}[1]{}

\title{\LARGE \bf
 Multimodal Belief-Space Covariance Steering \\with Active Probing and Influence for Interactive Driving
 }
\author{Devodita Chakravarty$^{1}$, John Dolan$^{2}$, Yiwei Lyu$^{3}$%
\thanks{$^{1}$Devodita Chakravarty is with the Department of Mechanical Engineering, 
Indian Institute of Technology Kharagpur, India. 
{\tt\small devoditac@kgpian.iitkgp.ac.in}}%
\thanks{$^{2}$John Dolan is with the Robotics Institute, 
Carnegie Mellon University, Pittsburgh, PA, USA. 
{\tt\small jdolan@andrew.cmu.edu}}%
\thanks{$^{3}$Yiwei Lyu is with the Department of Computer Science and Engineering, 
Texas A\&M University, College Station, TX, USA. 
{\tt\small yiweilyu@tamu.edu}}%
}

\begin{document}

\maketitle
\thispagestyle{empty}
\pagestyle{empty}

%%%%%%%%%%%%%%%%%%%%%%%%%%%%%%%%%%%%%%%%%%%%%%%%%%%%%%%%%%%%%%%%%%%%%%%%%%%%%%%%
\begin{abstract}

Autonomous driving in complex traffic requires reasoning under uncertainty. Common approaches rely on prediction-based planning or risk-aware control, but these are typically treated in isolation, limiting their ability to capture the coupled nature of action and inference in interactive settings. This gap becomes especially critical in uncertain scenarios, where simply reacting to predictions can lead to unsafe maneuvers or overly conservative behavior.
Our central insight is that safe interaction requires not only estimating human behavior but also shaping it when ambiguity poses risks. To this end, we introduce a hierarchical belief model that structures human behavior across coarse discrete intents and fine motion modes, updated via Bayesian inference for interpretable multi-resolution reasoning. On top of this, we develop an active probing strategy that identifies when multimodal ambiguity in human predictions may compromise safety and plans disambiguating actions that both reveal intent and gently steer human decisions toward safer outcomes. Finally, a runtime risk-evaluation layer based on Conditional Value-at-Risk (CVaR) ensures that all probing actions remain within human risk tolerance during influence.
Our simulations in lane-merging and unsignaled intersection scenarios demonstrate that our approach achieves higher success rates and shorter completion times compared to existing methods. These results highlight the benefit of coupling belief inference, probing, and risk monitoring, yielding a principled and interpretable framework for planning under uncertainty.

\end{abstract}

%%%%%%%%%%%%%%%%%%%%%%%%%%%%%%%%%%%%%%%%%%%%%%%%%%%%%%%%%%%%%%%%%%%%%%%%%%%%%%%%
% \vspace{-.4cm}
\section{Introduction}
%\yl{structure the introduction section following the logic in abstract: 1) a lot of works have been done in xx and xx to allow evs to reason and act under uncertainty. 2) however, they are usually treated in isolation. this becomes a problem if xxx, and xxx will xx, leading to xxx. 3) even for works that have both components, actions are mostly planned in a reactive manner as a passive response to the prediction/xxx. 4) this is could be unsafe if the prediction is multimodal and there is ambiguity in the prediction, and only plan for one or all possible outcomes could lead to either unsafe or overly conservative behavior. 5) what's even worse is the situations where there is ambiguity in the multimodal prediction (explain what this means, e.g., flat distribution over multiple possibilties...), and it's hard to use this info to inform robot action in a meaningful way. 6) speak out our philosophy: we believe capturing the interplay is important+active influence is a good way to go. 7) Then follow with contribution statements. Don't use bullet points, use one paragraph instead.}

Reasoning under uncertainty is essential for autonomous driving in complex traffic. Advances in probabilistic human behavior prediction and risk-aware control have enabled autonomous vehicles (AVs) to capture multimodal behaviors, estimate intent, and manage state distributions for both safety and performance \cite{chen2025hyperdimensional, candela2023risk, zheng2022belief, pilipovsky2023data}. However, prediction and control are often treated as separate, sequential modules, overlooking the fact that an AV’s actions can influence human drivers’ behaviors. In interactive traffic, where decisions are co-dependent, this lack of feedback is problematic \cite{sadigh2016information, hu2022active}.

Most existing methods remain reactive: they plan based on anticipated trajectories without actively reducing uncertainty or shaping human responses. While such strategies may suffice under low uncertainty, they become risky in multimodal situations where multiple futures are equally plausible \cite{gadginmath2025activeprobingmultimodalpredictions, wang2023active}. In these cases, the AV is left with either unsafe single-mode plans or overly conservative maneuvers that hedge against all possibilities \cite{ren2022chance, he2025active}. Treating uncertainty as a fixed external input, rather than as something shaped by interaction, leaves AVs vulnerable to indecision, inefficiency, and elevated risk in safety-critical situations.

We argue that safe and efficient interaction in uncertain traffic requires more than passive prediction: it demands active shaping of human behavior when ambiguity poses risks. Our key idea is to couple inference and action by equipping AVs with probing strategies that both reduce uncertainty and gently bias human decisions toward cooperative outcomes.

The \textbf{main contributions} of this paper are threefold: \textbf{(1)} we introduce a hierarchical belief model that captures human driving behavior across discrete intents and motion-level modes, updated via Bayesian inference to enable coarse-to-fine reasoning about multimodal interactions, providing interpretable high-level insights while retaining motion-level precision, \textbf{(2)} we explicitly account for the interplay between prediction and action by proposing probing strategies for the robot that not only reduce ambiguity but also strategically influence human decisions toward safer outcomes, and \textbf{(3)} we design a runtime risk-aware layer based on Conditional Value-at-Risk (CVaR) that complements uncertainty quantification by safeguarding against adverse tail events, ensuring probing actions remain safe and tolerable. Together, these components yield a purposeful interaction framework, with improved safety and enhanced efficiency demonstrated in scenarios like lane-merging and unsignaled intersections when compared to state-of-the-art baselines.

\section{Related Work}
\label{sec:relatedwork}
Predicting human behavior is fundamental to safe interaction in mixed-autonomy traffic. Unlike physical systems with unimodal dynamics, human driving decisions are inherently multimodal due to intent ambiguity (e.g., simultaneously considering yielding or merging aggressively). Capturing this uncertainty requires models that reason across both discrete intent and continuous motion levels. Bayesian inference provides effective online updates of such beliefs \cite{farhi2022bayesian}, and hierarchical structures have been proposed to couple these levels \cite{ludlow2024hierarchical, bao2025hierarchical}. This coarse-to-fine representation enables prediction of diverse futures while maintaining interpretability for decision-making. However, most prior work focuses on prediction alone, leaving open how to act effectively under multimodal beliefs.

Risk-aware planning frameworks address this by explicitly accounting for control uncertainty. Approaches include enforcing hard risk bounds, adding risk penalties to objectives, using chance constraints under probabilistic models, and adopting CVaR, which is distribution-free and penalizes adverse tail events \cite{yin2022risk, ahmadi2021risk}. Extensions to multimodal settings employ Gaussian mixtures \cite{ren2022chance} or game-theoretic formulations \cite{chandra2022game}, building on theoretical foundations in \cite{ren2022safe}. While effective in managing risk, these methods remain reactive: they assume fixed uncertainty distributions and do not actively reduce ambiguity. This becomes especially problematic when competing hypotheses have similar likelihoods, leaving the system indecisive at precisely the most safety-critical moments.

Belief-space planning further incorporates uncertainty into the state representation. Covariance steering offers analytical formulations for controlling both mean and uncertainty evolution with closed-form solutions for linear Gaussian systems \cite{liu2024optimal}. Extensions such as trajectory distribution control \cite{yin2022trajectory} and constrained formulations \cite{balci2025constrained, balci2022constrained} improve applicability. Belief-based human modeling with hierarchical intent–motion structures has also been explored \cite{ludlow2024hierarchical, bao2025hierarchical}. Yet these approaches primarily manage and propagate uncertainty rather than actively probing to reshape interaction.

The shift from passive observation to active information gathering has further shaped human-robot interaction. Early work in active perception focused on sensor control and information-theoretic planning \cite{elfes1991dynamic}. More recently, probing has been applied to interaction: balancing exploration and exploitation under dual control \cite{hu2022active}, uncovering hidden parameters \cite{wang2023active}, or influencing latent states through strategic actions \cite{sadigh2016information}. Probing has also been integrated with multimodal prediction \cite{gadginmath2025activeprobingmultimodalpredictions} and studied in adversarial contexts using game-theoretic models \cite{he2025active}. Despite these advances, most approaches neglect risk-awareness and rely on flat (non-hierarchical) belief representations, limiting their suitability for safety-critical driving.

Closest to our work, \cite{wang2023active} maximizes probing for behavior influence but relies on flat intent beliefs and lacks hierarchical modeling or risk-bounded probing. \cite{ren2022chance} applies CVaR-based risk-aware control under multimodal uncertainty but treats risk separately from intent inference, without actively reducing ambiguity. \cite{gadginmath2025activeprobingmultimodalpredictions} integrates probing with multimodal prediction but remains primarily reactive, without hierarchical intent structures or runtime risk monitoring. In contrast, our approach unifies hierarchical intent–motion belief modeling, active probing that both disambiguates uncertainty and influences human behavior, and CVaR-based runtime risk adjustment. This combination enables safe, interpretable, and efficient decision-making in uncertain interactive driving scenarios.
\vspace{-.3cm}
\section{Methodology}
\label{method}

Following our motivation, we aim to answer the following research questions:
\begin{enumerate}
% [label=Q\arabic*.]
    \item How should autonomous vehicles represent and update their beliefs about human intent and motion in uncertain, multimodal intents? 
    % \label{Q1}

    \item When ambiguity in human behavior predictions arises, how can it be identified and how can strategies be designed to actively disambiguate intent, so that autonomous vehicles can make safer and more informed decisions proactively rather than merely reacting?
    
    \item Consider risk in interactive decision-making: what is a more effective way to quantify and manage this risk without sacrificing efficiency, so that autonomous decisions remain safe and responsive in uncertain, dynamic traffic scenarios?
    \label{Q2}

    \label{Q3}
    
    \label{Q4}
\end{enumerate}

\noindent To address Q1, Sec.~\ref{sec:hierarchical_belief} introduces a hierarchical belief model for human behavior prediction. For Q2, Sec.~\ref{sec:active_probing} presents active probing to reveal hidden information. For Q3, Sec.~\ref{sec:multimodal_belief} details risk quantification from multimodal belief, while Sec.~\ref{sec:covariance_steering} develops covariance steering to actively shape human behavior and its associated risk. Finally, Sec.~\ref{sec:feedback_loop} brings these elements together through a bi-directional feedback loop.

\vspace{-0.39 cm}
\subsection{Human Behavior Prediction with Hierarchical Belief Model}
\label{sec:hierarchical_belief}

Let the ego agent's state at time $t$ be denoted by $x(t) \in \mathbb{R}^n$, 
and the observation of a surrounding agent be $\bm{z}(t)$. We model the latent 
intent of the surrounding agent as a discrete random variable $i \in \{1, \ldots, I\}$. 
These \textbf{intents} represent high-level, mutually exclusive behavioral goals, 
such as \textit{yielding}, \textit{merging aggressively}, or \textit{cruising neutrally}. 
Our belief over these intents is captured by the probability vector 
$\pi(t) = [\pi_1(t),\ldots,\pi_I(t)]^T$.

This high-level representation, however, is too coarse to capture the nuances of 
agent behavior. To address this, each intent $i$ is further refined by a set of 
behavioral \textbf{modes} $k \in \{1,\ldots,K_i\}$, with associated weights 
$w_{i,k}(t)$. These modes represent different ways an intent can be realized; for 
example, a \textit{yielding} intent can be executed via a \textit{gradual deceleration} 
mode or a \textit{sudden braking} mode. Each mode $(i,k)$ is formally defined by 
a target Gaussian distribution $\mathcal{N}(\mu_{i,k}, \Sigma_{i,k})$, where 
$\mu_{i,k}$ encodes the nominal state that behavior aims to achieve and 
$\Sigma_{i,k}$ captures the variability around it. These parameters are specified 
offline (e.g., via clustering of demonstrations or expert priors~\cite{de2021dynamic}) 
and remain fixed during online inference. These Gaussian parameters serve two roles: 
they define the likelihood model used in the Bayesian belief update (Eq.~1--3) and 
specify the target mean and covariance for the covariance-steering controller (Eq.~7).

To address \textbf{Q1},
% \textbf{\ref{Q1}} \yl{fix this reference issue}, 
we maintain a hierarchical belief over discrete intents 
and their associated motion modes. The process begins at the lowest level: given a 
new observation $\bm{z}(t)$, we compute the likelihood of each mode $(i,k)$: 
$\ell_{i,k}(t) = p(\bm{z}(t) \mid i, k)$, which serves as the atomic piece of 
evidence grounding our inference in the most recent sensor data.

\begin{algorithm}[t]
\footnotesize
\caption{Hierarchical Belief Update}
\label{alg:belief_update}
\begin{algorithmic}[1]
\STATE \textbf{Input:} Observation $\bm{z}(t)$, prior beliefs $\pi(t), \{w_{i,k}(t)\}$
\STATE \textbf{Output:} Updated beliefs $\pi(t+1), \{w_{i,k}(t+1)\}$
\FOR{$i = 1$ to $I$}
    \FOR{$k = 1$ to $K_i$}
        \STATE $\ell_{i,k}(t) \leftarrow p(\bm{z}(t)\mid i,k)$
    \ENDFOR
\ENDFOR
\FOR{$i = 1$ to $I$}
    \STATE $\mathcal{L}[i] \leftarrow \sum_{k=1}^{K_i} w_{i,k}(t)\,\ell_{i,k}(t)$
\ENDFOR
\STATE $Z \leftarrow \sum_{j=1}^{I} \pi_j(t)\,\mathcal{L}[j]$
\FOR{$i = 1$ to $I$}
    \STATE $\pi_i(t+1) \leftarrow \dfrac{\pi_i(t)\,\mathcal{L}[i]}{Z}$
    \STATE $Z_i \leftarrow \sum_{j=1}^{K_i} w_{i,j}(t)\,\ell_{i,j}(t)$
    \FOR{$k = 1$ to $K_i$}
        \STATE $w_{i,k}(t+1) \leftarrow \dfrac{w_{i,k}(t)\,\ell_{i,k}(t)}{Z_i}$
    \ENDFOR
\ENDFOR
\STATE \textbf{return} $\pi(t+1), \{w_{i,k}(t+1)\}$
\end{algorithmic}
\end{algorithm}

With evidence for each individual mode, we ascend the hierarchy to the intent level. 
The aggregated likelihood for intent $i$ is a weighted sum over its constituent modes:
\begin{equation}\footnotesize
    \mathcal{L}[i] = \sum_{k=1}^{K_i} w_{i,k}(t)\,\ell_{i,k}(t).
\end{equation}
This ensures that intent likelihood reflects the entire distribution of associated 
behaviors, not just the single best-fitting mode. Modes that were already considered 
probable ($w_{i,k}(t)$ is high) contribute more significantly, creating natural 
temporal smoothing of the belief.

To combine aggregated likelihoods with our prior $\pi(t)$, we apply Bayes' rule. 
The normalization constant $Z = \sum_{j=1}^I \pi_j(t)\,\mathcal{L}[j]$ represents 
the total probability of observing $\bm{z}(t)$ across all intents and modes. 
The posterior intent probability is then:
\begin{equation}\footnotesize
    \pi_i(t+1) = \frac{\pi_i(t)\,\mathcal{L}[i]}{Z}, \quad \forall i \in \{1, \ldots, I\}.
\end{equation}
Intents consistent with the new observation have their probabilities amplified, 
while inconsistent ones are suppressed.

Finally, mode weights \textit{within} each intent are updated analogously:
\begin{equation}\footnotesize
    w_{i,k}(t+1) = \frac{w_{i,k}(t)\,\ell_{i,k}(t)}{\sum_{j=1}^{K_i} 
    w_{i,j}(t)\,\ell_{i,j}(t)}.
\end{equation}
This is effectively a separate Bayesian update performed conditionally for each 
intent $i$, redistributing probability mass toward modes that better explain 
$\bm{z}(t)$.

Together, these coupled updates (Eq.~2 and~3) sharpen beliefs at both levels 
simultaneously, preventing incoherence where a mode remains highly likely 
($w_{i,k}$ high) even though its parent intent has become negligible 
($\pi_i \approx 0$). The full procedure is summarized in 
Algorithm~\ref{alg:belief_update}.
% \vspace{-.2cm}
\subsection{Reveal Hidden Information with Active Probing}
\label{sec:active_probing}

In uncertain traffic scenarios, ambiguity about human intent and motion can compromise both safety and efficiency. Addressing \textbf{Q2}, which asks how ambiguity can be identified and disambiguated, we introduce an \textbf{active probing} strategy that deliberately seeks information to clarify uncertainty. Instead of passively reacting to predictions, the ego vehicle selects actions that reduce the uncertainty in future beliefs by exploring ambiguous scenarios.

The probing objective is based on \textbf{entropy}, which measures uncertainty in the belief distributions over intents and motion modes. The entropy at time $t$ is computed as:
\begin{equation}\footnotesize
\begin{split}
H(\pi(t), w(t)) &= - \sum_{i=1}^I \pi_i(t) \log \pi_i(t) \\
&\quad - \sum_{i=1}^I \sum_{k=1}^{K_i} w_{i,k}(t) \log w_{i,k}(t)
\end{split}
\end{equation}
where $\pi_i(t)$ is the belief over intent $i$ and $w_{i,k}(t)$ is the belief over mode $k$ within intent $i$.

To encourage actions that actively gather information, we define the probing objective as the expected reduction in uncertainty:
\begin{equation}\footnotesize
    J_{\text{probe}}(u) = -\lambda_H \, \mathbb{E}[H(\pi(t+1), w(t+1))],
\end{equation}
where $\lambda_H > 0$ scales the importance of probing, and the expectation is taken over possible future observations that result from applying control $u$. By minimizing this objective, the ego vehicle selects actions that are expected to reveal hidden aspects of human behavior, making subsequent interaction safer and more informed.

% \vspace{-.2cm}
\subsection{Uncertainty Quantification with Multimodal Belief}
\label{sec:multimodal_belief}

For \textbf{Q3}, which concerns how to effectively quantify and manage risk without sacrificing efficiency, we formalize the ego’s uncertainty using a hierarchical belief structure. This structure explicitly represents the uncertainty across multiple behavioral modes and intents.

The belief over intents at time $t$ is expressed as: $\pi(t) = [\pi_1(t), \dots, \pi_I(t)]$,
where $\pi_i(t)$ represents the probability of intent $i$. For each intent, the belief over motion modes is given by: $w(t) = \{w_{i,k}(t)\},  i = 1, \dots, I,  k = 1, \dots, K_i$,
where $w_{i,k}(t)$ represents the probability of mode $k$ for intent $i$. Combining these, the joint probability of the agent following mode $(i,k)$ is: $\omega_{i,k}(t) = \pi_i(t+1) w_{i,k}(t+1)$,
which reflects the agent’s overall behavior under the updated belief.

To account for safety under deep uncertainty, we incorporate \textbf{Conditional Value-at-Risk (CVaR)}, which focuses on the worst-case scenarios rather than the average behavior. Given a set of $S$ simulated future trajectories with associated costs $J^{(s)}_{i,k}$, the CVaR at confidence level $\alpha$ is defined as:
\begin{equation} \footnotesize
    \text{CVaR}_\alpha(i,k) = \frac{1}{m} \sum_{s=1}^m J^{(s)\uparrow}_{i,k}, \quad m = \max(1, \lceil \alpha S \rceil),
\end{equation}
where $J^{(s)\uparrow}_{i,k}$ are the sorted costs in descending order and $\alpha \in (0,1)$ is a chosen risk tolerance parameter. This approach allows the ego vehicle to balance both likelihood and potential danger, thereby quantifying uncertainty in a manner that explicitly integrates safety and efficiency considerations.
% \vspace{-.2cm}
\subsection{Actively Shaping Human Belief via Covariance Steering}
\label{sec:covariance_steering}

To actively shape human behavior while ensuring safety and efficiency, we address both \textbf{Q2} and \textbf{Q3} by formulating a finite-horizon covariance steering optimization problem over horizon T. This framework enables the ego vehicle to guide its own state trajectory toward desired outcomes while controlling the associated uncertainty distribution by explicitly considering both expected behavior and worst-case risk.
For each mode $(i,k)$, the target behavior, which is the velocity gaussian of the ego's belief on human intent,
%\yl{what does target behavior mean? A behavior is a gaussian?} 
is specified by a Gaussian distribution $\mathcal{N}(\mu_{i,k}, \Sigma_{i,k})$,
where $\mu_{i,k}$ encodes the desired state and $\Sigma_{i,k}$ encodes the allowable variability. These targets are determined offline through expert knowledge, clustering of observed trajectories, or prior experience \cite{de2021dynamic} and reflect safe and cooperative behavior patterns. The ego’s predicted state at time $t$ is modeled by $\mathcal{N}(\mu(t), \Sigma(t))$, 
where $\mu(t)$ and $\Sigma(t)$ represent the predicted mean and covariance at time $t$ given the applied control sequence.

The ego vehicle’s objective is to minimize both the expected uncertainty and the deviation from desired behavior. This is compactly expressed using the previously defined objectives:
\begin{equation}\footnotesize
    u^* = argmin_u \; J(u) = J_{\text{probe}}(u) + J_{\text{influence}}(u),
\end{equation}
where $J_{\text{probe}}(u)$ captures the information gain through active probing (Eq.~5) and $J_{\text{influence}}(u)$ quantifies alignment with the target behavior using the Kullback-Leibler divergence:
\begin{equation}\footnotesize
    J_{\text{influence}}(u) = D_{\text{KL}}\big(\mathcal{N}(\mu(t), \Sigma(t)) \,\|\, \mathcal{N}(\mu_{i,k}, \Sigma_{i,k})\big).
\end{equation}

The ego’s state evolves according to discrete-time dynamics \cite{okamoto2019optimal}:
\begin{equation}\footnotesize
    \mu(t+\tau+1) = A\mu(t+\tau) + Bu(t+\tau),
\end{equation}
\begin{equation}\footnotesize
    \Sigma(t+\tau+1) = A\Sigma(t+\tau)A^\top + BWB^\top,
\end{equation}
where $A$ and $B$ are linear approximations of the system dynamics, and $W$ is the process noise covariance matrix.

To explicitly ensure robustness against worst-case scenarios, we impose Conditional Value-at-Risk (CVaR) constraints for each mode:
\begin{equation}\footnotesize
    \text{CVaR}_\alpha(i,k) \leq \bar{J},
\end{equation}
where $\bar{J}$ is a predefined risk threshold and $\alpha \in (0,1)$ is the confidence level for tail risk. The control inputs are constrained to remain within safe operational bounds: $u_{\min} \leq u(t+\tau) \leq u_{\max}, \quad \forall \tau = 0,\dots,T-1$,
ensuring that the vehicle operates within actuator limits and avoids unsafe maneuvers. State-of-the-art quadratic programming solvers such as OSQP (Operator Splitting Quadratic Program) can be used to solve this optimization problem. Even though this formulation offers a mathematically sound method with precise goals and limitations, it could be computationally costly to solve the entire optimization at each time step. Approximate techniques with warm-starts, like sequential quadratic programming or model predictive control (see Sec \ref{experiment}), can be used in real-world applications. Furthermore, solutions can be improved in between solver iterations using gradient-based updates, guaranteeing responsiveness in real-time interactions  \cite{stellato2020osqp, hall2022real}.
By explicitly defining this covariance steering optimization, we create a principled framework that integrates safety, efficiency, and information gain while allowing for solver-based implementation that can be tailored to real-time operational requirements.

% \vspace{-.2cm}
\subsection{Bi-Directional Prediction-Action Feedback Loop}
\label{sec:feedback_loop}

\begin{algorithm}[t]
\footnotesize
\caption{Active Multimodal Covariance Steering with Risk-aware Probing}
\label{alg:combined_cov}
\begin{algorithmic}[1]
\STATE \textbf{Input:} Initial state $x(0)$, target distributions $\{\mu_{i,k}, \Sigma_{i,k}\}$, horizon $T$
\STATE \textbf{Output:} Trajectory $\{x(t)\}_{t=0}^{T}$

\STATE Initialize beliefs: $\pi_i(0) \leftarrow 1/I$, $w_{i,k}(0) \leftarrow 1/K_i$

\FOR{$t = 0$ to $T-1$}
    \STATE Observe environment: $\bm{z}(t)$
    \STATE Update beliefs: $(\pi(t+1), w(t+1)) \leftarrow \textsc{BeliefUpdate}(\bm{z}(t), \pi(t), w(t))$
    \STATE Compute weighted beliefs: $\omega_{i,k} \leftarrow \pi_i(t+1)\, w_{i,k}(t+1)$
    \STATE Estimate CVaR: $\text{CVaR}_\alpha(i,k)$ for each mode
    \STATE $u_{\text{total}} \leftarrow 0$

    \FOR{all $(i,k)$ such that $\omega_{i,k} > \epsilon$}
        \STATE Formulate probing objective $J_{\text{probe}}(u)$
        \STATE Formulate influence objective $J_{\text{influence}}(u)$
        \STATE $J(u) \leftarrow J_{\text{probe}}(u) + J_{\text{influence}}(u)$
        \STATE $\nu_{i,k}(t) \leftarrow \arg\min_u J(u)$
        \STATE \hspace{0.5em} subject to $\{\mu_{i,k}, \Sigma_{i,k}\}$,
        \STATE \hspace{0.5em} $\text{CVaR}_\alpha(i,k) \leq \bar{J}$,
        \STATE \hspace{0.5em} $u_{\min} \leq u \leq u_{\max}$
        \STATE $u_{\text{total}} \leftarrow u_{\text{total}} + \omega_{i,k}\,\nu_{i,k}(t)$
    \ENDFOR

    \STATE Apply control: $x(t+1) \leftarrow f(x(t), u_{\text{total}}, \Delta t)$
    \IF{SafetyViolation$(x(t+1))$}
        \STATE \textbf{break}
    \ENDIF
\ENDFOR

\STATE \textbf{return} $\{x(t)\}_{t=0}^{T}$
\end{algorithmic}
\end{algorithm}

A closely coupled feedback loop 
that incorporates belief updates, probing, and covariance steering makes it possible for prediction and action to interact. Every time step, the following sequence is carried out by the ego vehicle:
\begin{enumerate}
    \item \textbf{Perceive:} The ego gathers a fresh observation z(t) from its sensors, capturing the motion of agents in the immediate vicinity.
    \item \textbf{Infer:} The ego propagates uncertainty across both high-level intents and motion modes by using the Hierarchical Belief Update to refine its belief distributions $\pi(t)$ and $w_{i,k}(t)$ in light of the observation.
    \item \textbf{Decide \& Act:} 
% The risk-conscious 
By minimizing the objective $J(u)$, the probing controller integrates the influence and probing objectives to calculate the control action $u(t)$. This choice aligns the ego's trajectory towards safe and cooperative behaviors, takes into consideration the current uncertainty, and gives information acquisition priority in ambiguous situations.
    \item \textbf{Propagate:} The ego’s state is updated as: $x(t+1) \leftarrow f(x(t), u(t), \Delta t)$,
    where the updated state depends on the applied control and time step.
\end{enumerate}

Algorithm 2 summarizes this bi-directional feedback loop, where the ego’s actions continuously reshape its beliefs, which in turn guide subsequent actions. The ego actively seeks information to clarify intent when entropy is high; when uncertainty decreases, it shifts to efficient execution, focusing on cooperative actions that are consistent with low-risk pathways. The ego balances exploration and exploitation in real time by dynamically modifying its trajectory and related uncertainty through covariance steering.  The resulting interactions provide a practical and consequence-aware approach to uncertainty-aware planning in complex traffic situations, consistent with human-driven behaviors, safer, and easier to interpret.

% \vspace{-.3cm}
\section{Simulation \& Discussion}
\label{sec:results}

\begin{figure*}[t]
    \centering
    \includegraphics[width=1.00\textwidth,height=\textheight,keepaspectratio]{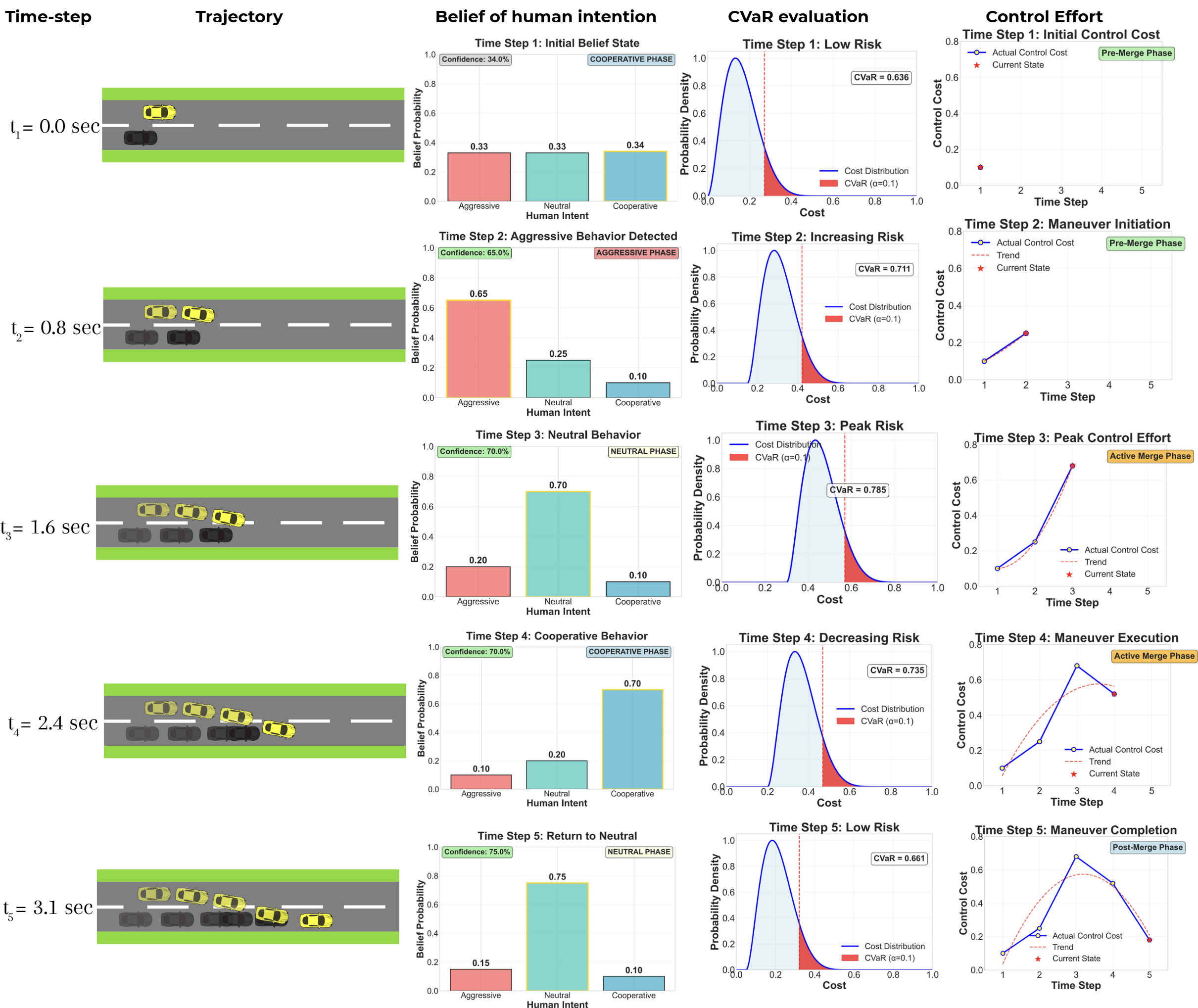}
    \caption{\footnotesize Overview of the lane-merging interaction between an autonomous ego vehicle and a human-driven vehicle, illustrating the evolution of hierarchical belief updates, CVaR-based risk assessment, control effort, and resulting trajectories from $t_1$ to $t_5$. The progression shows how active probing and Bayesian updates refine intent inference from uncertain to cooperative, while control effort adapts to balance safety and efficiency. Specifically, the \textit{hierarchical belief update} displays Bayesian updates integrating noisy observations and probing-induced responses, evolving from a uniform intent distribution at $t_{1}$ to a concentrated cooperative belief at $t_{5}$. Instead of passively observing, the \textit{active probing controller} at $t_{2}$ demonstrates how probing actions intentionally change the interaction and belief state to reveal latent human intentions. The \textit{CVaR risk assessment} curve illustrates how risk-aware control adjusts the vehicle's aggressiveness and conservatism during the interaction by first rising and then falling over time steps. Lastly, the model's ability to dynamically combine efficiency and safety is demonstrated by the way the \textit{adaptive control effort} changes in response to inferred risk and intent certainty.}
%    \caption{\footnotesize Overview of our proposed framework for risk-aware multimodal interaction under uncertainty in a lane merge scenario. Columns show the trajectory, evolving belief over human intent, CVaR-based risk evaluation, and control effort across time steps. The sequence highlights how probing refines beliefs, manages risk, and shapes human responses. }
% \caption{\footnotesize Overview of the lane-merging interaction between an autonomous 
% ego vehicle and a human-driven vehicle, illustrating the evolution of hierarchical 
% belief updates, CVaR-based risk assessment, control effort, and resulting trajectories 
% from $t_1$ to $t_5$. The \textit{hierarchical belief update} displays Bayesian updates 
% integrating noisy observations and probing-induced responses, evolving from a uniform 
% intent distribution at $t_{1}$ to a concentrated cooperative belief at $t_{5}$. The 
% \textit{active probing controller} at $t_{2}$ demonstrates how probing actions 
% intentionally reshape the interaction and belief state to reveal latent human 
% intentions. The \textit{CVaR risk assessment} curve illustrates how risk-aware control 
% adjusts the vehicle's aggressiveness and conservatism, first rising then falling across 
% time steps. The \textit{adaptive control effort} correspondingly shifts in response to 
% inferred risk and intent certainty, reflecting the dynamic balance between safety and 
% efficiency.}
    \label{fig:method}
\end{figure*}

% In this section, we present an in-depth analysis of our proposed framework through simulational evaluations IV-B \& C, followed by a comparative study against established baselines \ref{comp}.

% \yl{organize the simulation and discussion section as follows:
% 1) simulation setup: introduce two scenarios, and the parameters you use;
% 2) validity of the approach: by showing the successful rate and/or other metrics from a large number of trials of experiments (don't forget to define what is considered sucessful, success in merging or sucess in steering the belief/probe...); 
% 3) example case studies to better illustrate the mechanism of our approach (Fig. 1 and 2, although the figure 2 is only showing the scenario plot, the description of the key information in the example should be provided);
% 4) Comparative analysis with baselines (this is where you introduce other methods, and report their performance in the same set of tests you gave earlier to our own approach. and explain why the diference.
% }

\subsection{Experimental Setup}
\label{experiment}

We evaluate the framework in two interactive driving scenarios: lane merging and an unsignaled four-way intersection, chosen for their challenges in longitudinal merging and lateral conflict resolution. Human-driven agents are modeled deterministically, reacting predictably to the ego vehicle’s probing. The human model follows \cite{siebinga2024model}, which generates best-response plans updated by risk thresholds and implicit communication. We adapt it as a reactive trajectory generator within our hierarchical belief structure, allowing integration with Bayesian updates and active probing so the ego can refine beliefs and influence human responses toward safe, cooperative outcomes.

%\yl{how many trials are the experiments for each of the scenario? Don't put comma between values and units. In this part, there should be two components: 1) the statistical results from the large numbers of trials for quantitative analysis (currently missing), and 2) two example trials to demonstrate what happened for qualitative analysis.}
Both the scenarios are run for 500 runs each. Initial positions and velocities of all vehicles are sampled uniformly. In lane merges, ego velocity is between $7$–$10 \text{ m/s}$, with surrounding traffic between $8$–$12 \text{ m/s}$. In intersections, ego velocity is $4$–$6 \text{ m/s}$ and the crossing agent $5$–$8 \text{ m/s}$. Vehicle positions are randomized around the merge point or intersection center. Horizons are $T=4$ s (merge) and $T=6$ s (intersection) with a $10$ Hz rate. Risk-sensitivity is incorporated via CVaR regularization, with $\alpha \in [0.05, 0.2]$ to balance tail-risk and efficiency. Across runs, best-performing parameters were $S=100$, $\alpha=0.05$, $\beta=5$, $\lambda_H=0.5$, $\gamma=0.95$, $\epsilon=10^{-5}$, horizon length of $30$ steps ($\Delta t=0.1$, corresponding to $T=3$ s), and $K_i=3$ intent modes. We solved the optimization by solving a Model Predictive Control (MPC) at each time step. Here, the problem is repeatedly solved over a finite horizon with updated state estimates, and only the first control input is applied 
% \clearpage
before the optimization is resolved at the next time step.

% \vspace{-.2cm}
\subsection{Quantitative Analysis of Proposed Method}

In order to evaluate the effectiveness of our proposed approach, we considered key performance metrics that reflect both safety and efficiency in interactive driving scenarios. These include \textit{Success Rate}, which measures the percentage of simulation runs where the ego vehicle safely completes the maneuver; \textit{Time to Merge} or \textit{Time to Cross}, which indicates how quickly interactions are resolved; \textit{Gap to Vehicle}, representing the distance maintained from surrounding agents; and \textit{Velocity}, which reflects how efficiently the ego vehicle moves toward its objective. Additionally, \textit{Longitudinal} and \textit{Angular Jerk} are used to quantify the smoothness of maneuvers, ensuring that rapid or uncomfortable control changes are avoided. These metrics collectively capture how well the ego vehicle balances safety, efficiency, and comfort under uncertainty.

As shown in Table I and Table II (last columns), our method consistently outperforms the baseline approaches across both the lane merging and unsignaled intersection scenarios. For example, we achieved a success rate of \textbf{96\%} in lane merging and \textbf{94\%} in intersections, demonstrating the robustness of our hierarchical belief model combined with risk-aware probing. The reduction in \textit{Time to Merge} (to \textbf{2.03 ± 1.38 s}) and \textit{Time to Cross} (to \textbf{4.14 ± 1.87 s}) shows that our method enables faster decision-making without compromising safety. 
The improvements in \textit{Gap to Vehicle} and \textit{Velocity} reflect how the ego vehicle can safely operate closer to other agents and at higher speeds, leveraging the CVaR-based tail-risk adjustment to ensure that such actions remain within acceptable safety bounds. Furthermore, the comparable or improved values for \textit{Longitudinal} and \textit{Angular Jerk} indicate that these efficiency gains do not result in abrupt or uncomfortable maneuvers. These results highlight how actively shaping uncertainty through probing and influence, rather than passively reacting, enables safer, smoother, and more efficient interactions in complex traffic environments.
% \vspace{-.3cm}
\subsection{Example Case Studies}

Fig. 1 
%\yl{missing fig ref} 
presents a step-by-step illustration of the autonomous vehicle (ego) interacting with a human-driven vehicle during a lane merging maneuver. The figure visualizes the progression of the hierarchical belief update, risk assessment, control effort, and resulting trajectories across time steps $t_{1}$ through $t_{5}$.  At $t_{1}$ (Uncertain Initial Belief), the ego vehicle starts with a prior hierarchical belief over the human driver’s latent intent, distributed over the three discrete intent categories: aggressive, neutral, and cooperative. This belief is represented probabilistically as the vector $\pi(t_{1}) = [\pi_{\text{aggressive}}, \pi_{\text{neutral}}, \pi_{\text{cooperative}}](t_{1})$, reflecting the initial uncertainty. Each intent further decomposes into multiple motion modes $k$, each with corresponding Gaussian targets $\mathcal{N}(\mu_{i,k}, \Sigma_{i,k})$. The prior belief is broad due to limited observation at this early stage, reflecting ambiguity about how the human will behave. At $t_{2}$ (Active Probing Increases Aggressive Intent Likelihood), the ego executes an active probing maneuver designed to reduce uncertainty by influencing the human’s behavior and eliciting informative responses. This maneuver is computed by minimizing the expected future entropy of the hierarchical belief and steering the system to maximize information gain about human intent (Eq.~4, 5). As a result, after observing the human response to this probing action, the Bayesian belief update increases the probability weight on the aggressive intent mode, $\pi_{\text{aggressive}}(t_{2})$. Simultaneously, the tail risk of collision or unsafe interaction rises, as measured by the Conditional Value-at-Risk (CVaR) metric, since aggressive behavior carries higher risk in the merge context. Consequently, the control effort by the ego vehicle increases at $t_{2}$, as it adapts its maneuvers to safely accommodate the possibility of aggressive human actions while maintaining efficiency constraints. At $t_{3}$ (Human Response Shifts Belief Toward Neutral), the human driver reacts to the ego’s probing by adjusting their motion (velocity), which is observed and incorporated in the belief update. This results in shifting posterior belief from aggressive towards the neutral intent mode, $\pi_{\text{neutral}}(t_{3})$, reflecting reduced risk. The CVaR risk estimation correspondingly decreases as the system’s uncertainty clarifies and less adverse outcomes are forecasted. The ego’s control effort adjusts downward, balancing safety with the new, safer predicted human behavior. At $t_{4}$ (Human Yields; Cooperative Intent Becomes Most Likely), further interaction by probing and continuous Bayesian updates refines the belief and influences the human’s intention to yield, reflected as increased probability on the cooperative intent mode. This cooperative mode corresponds to trajectories with smoother, less adversarial human motions that facilitate a safe merge. The reduction in CVaR risk continues, as the threat of collision or emergency braking diminishes. The ego vehicle exploits this clarity to optimize for smoother, more efficient control inputs. At $t_{5}$ (Maneuver Completion with Reduced Control Effort), the ego confidently completes the lane merge after the human's intent has been sufficiently resolved and cooperative behavior verified. There is little remaining uncertainty indicated by the low entropy of the hierarchical belief distribution, which is centred around the cooperative mode. A safe interaction envelope is reflected in the low CVaR-driven tail-risk measure. The lowest level of control efforts is a sign of effective, well-thought-out execution made possible by earlier active probing and risk-conscious planning.

% The Fig. 1 demonstrates how each component contributes to the overall goal. The \textit{hierarchical belief update} displays Bayesian updates integrating noisy observations and probing-induced responses, evolving from a uniform intent distribution at $t_{1}$ to a concentrated cooperative belief at $t_{5}$. Instead of passively observing, the \textit{active probing controller} at $t_{2}$ demonstrates how probing actions intentionally change the interaction and belief state to reveal latent human intentions. The \textit{CVaR risk assessment} curve illustrates how risk-aware control adjusts the vehicle's aggressiveness and conservatism during the interaction by first rising and then falling over time steps. Lastly, the model's ability to dynamically combine efficiency and safety is demonstrated by the way the \textit{adaptive control effort} changes in response to inferred risk and intent certainty.

% Our claim that combining tail-risk-aware control, active exploration, and hierarchical intent inference results in safer, easier-to-understand, and more effective manoeuvres in interactive driving situations is supported by this thorough sequential visualisation.

\begin{figure}[h]
\centering
\includegraphics[width=.4\linewidth]{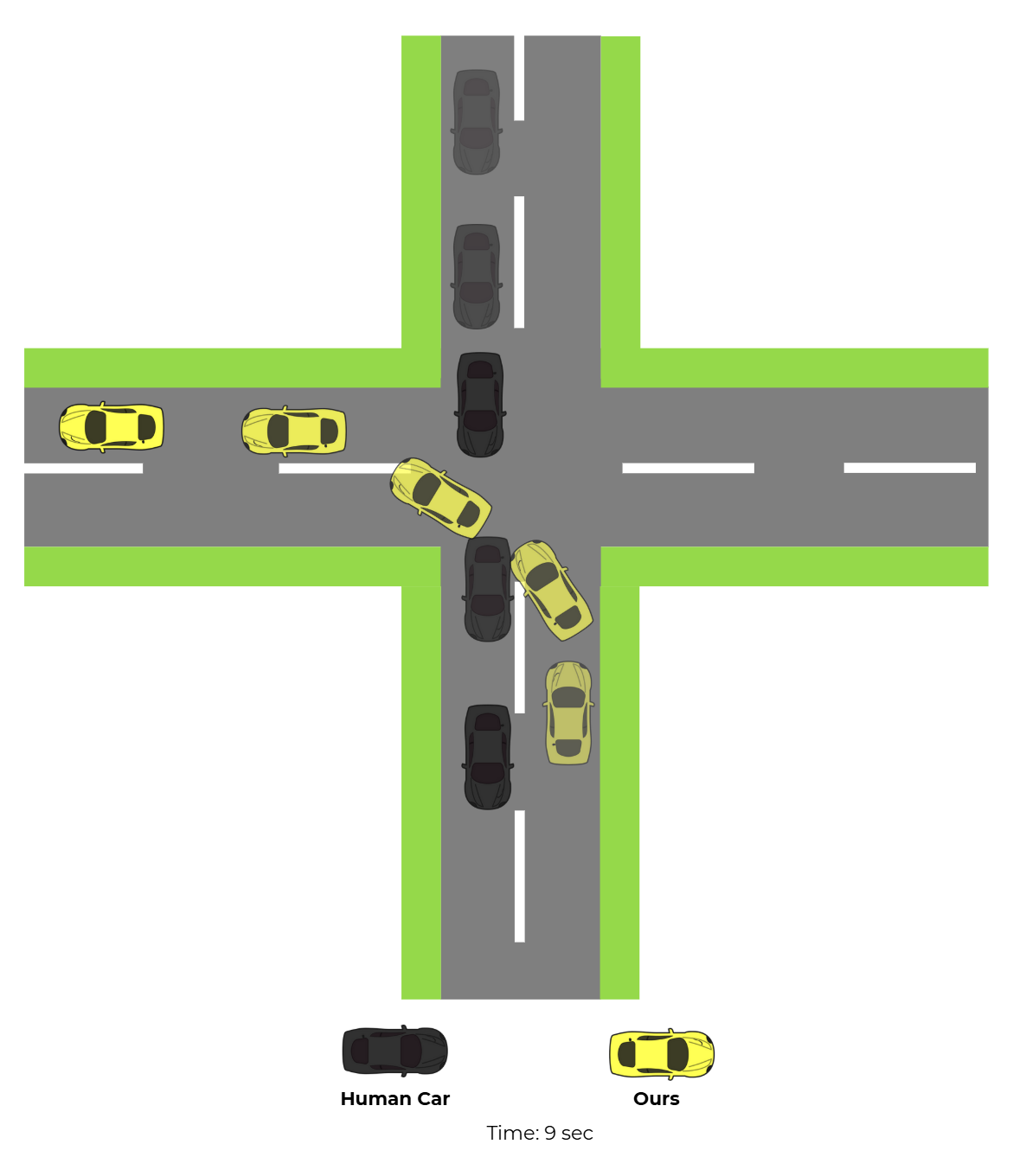}
\caption{\footnotesize Unsignaled four-way intersection. The ego shapes the crossing agent’s belief, leading it to yield.}
\label{fig:intersection}
\end{figure} 
% \vspace{-.5cm}
The intersection scenario (Fig.~\ref{fig:intersection}) further highlights probing as signaling. Both vehicles arrive simultaneously with no right of way. The ego performs small, risk-bounded probes that are interpreted as a commitment to cross first, prompting the human to yield. This avoids deadlock and resolves conflict while keeping actions within CVaR limits.

\subsection{Comparative Analysis with Baselines}
\label{comp}
\begin{table*}[t]
\centering
\footnotesize
\caption{Comparison of Performance Metrics for lane merging scenario for 500 simulation runs}
\label{lane}
\begin{tabular}{lcccc}
\toprule
\textbf{Metric} & \textbf{AP-IH} & \textbf{CC-MPC} & \textbf{AP-MP} & \textbf{Ours} \\
\midrule
Success Rate & $82\%$ & $87\%$ & $92\%$ & $\mathbf{96\%}$ \\
Time to Merge (s) & $6.80 \pm 1.34$ & $4.72 \pm 1.65$ & $5.25 \pm 1.42$ & $\mathbf{2.03 \pm 1.38}$ \\
Gap to Vehicle (m) & $9.54 \pm 3.60$ & $9.84 \pm 2.10$ & $8.10 \pm 2.20$ & $\mathbf{7.60 \pm 2.29}$ \\
Velocity (m/s) & $8.35 \pm 1.06$ & $7.90 \pm 1.08$ & $8.01 \pm 1.17$ & $\mathbf{8.88 \pm 1.02}$ \\
Longitudinal Jerk (m/s$^3$) & $0.31 \pm 0.18$ & $0.28 \pm 0.19$ & $\mathbf{0.21 \pm 0.22}$ & $0.22 \pm 0.16$ \\
Angular Jerk (rad/s$^3$) & $(1.52 \pm 1.21)\!\times\!10^{-2}$ & $(1.84 \pm 1.10)\!\times\!10^{-2}$ & $(1.76 \pm 1.05)\!\times\!10^{-2}$ & $\mathbf{(1.49 \pm 0.98)\!\times\!10^{-2}}$ \\
\bottomrule
\end{tabular}
\end{table*}
\begin{table*}[t]
\centering
\footnotesize
\caption{Comparison of Performance Metrics for an unsignaled four-way intersection scenario for 500 simulation runs}
\label{4way}
\begin{tabular}{lcccc}
\toprule
\textbf{Metric} & \textbf{AP-IH} & \textbf{CC-MPC} & \textbf{AP-MP} & \textbf{Ours} \\
\midrule
Success Rate & $78\%$ & $84\%$ & $88\%$ & $\mathbf{94\%}$ \\
Time to Cross (s) & $8.02 \pm 2.04$ & $5.78 \pm 1.90$ & $6.80 \pm 2.90$ & $\mathbf{4.14 \pm 1.87}$ \\
Gap to Vehicle (m) & $6.20 \pm 2.10$ & $6.80 \pm 2.00$ & $6.10 \pm 1.90$ & $\mathbf{5.70 \pm 2.00}$ \\
Velocity (m/s) & $4.56 \pm 0.78$ & $4.73 \pm 0.72$ & $4.84 \pm 0.67$ & $\mathbf{5.11 \pm 0.61}$ \\
Longitudinal Jerk (m/s$^3$) & $0.21 \pm 0.15$ & $0.20 \pm 0.16$ & $0.19 \pm 0.14$ & $\mathbf{0.16 \pm 0.13}$ \\
Angular Jerk (rad/s$^3$) & $0.18 \pm 0.12$ & $0.15 \pm 0.11$ & $\mathbf{0.12 \pm 0.10}$ & $0.14 \pm 0.11$ \\
\bottomrule
\end{tabular}
\end{table*}
\begin{figure*}[t]
    \centering
\includegraphics[width=1\textwidth]{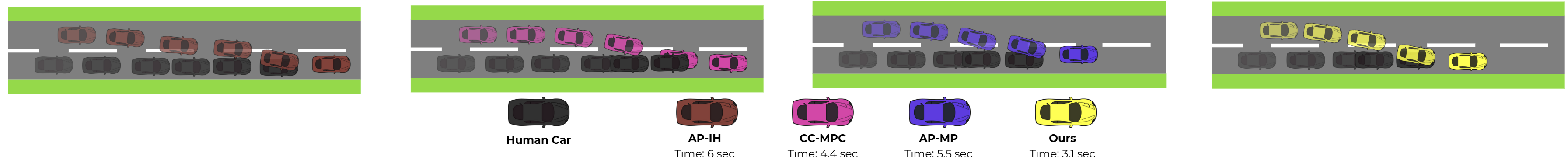}
\caption{\footnotesize Trajectory comparison across methods for the same lane merge scenario. AP-IH (brown), CC-MPC (pink), and AP-MP (blue) manage to merge but require longer times or more conservative maneuvers, while our method (yellow) merges more quickly and with smoother motion, reflecting the efficiency and risk-aware confidence of our planner.}    \label{fig:trajectory_comparison}
\end{figure*}
We compare our approach against three representative baselines from the literature: AP-IH \cite{wang2023active}, which explicitly formulates an information-maximizing probing objective to reveal human-model parameters and then uses the gathered information to influence behavior; CC-MPC \cite{ren2022chance}, a state-of-the-art trajectory planner that handles multimodal obstacle uncertainty using Gaussian mixture models and chance/CVaR approximations; and AP-MP \cite{gadginmath2025activeprobingmultimodalpredictions}, which couples multimodal prediction with an active probing strategy to reduce prediction ambiguity during planning. These baselines were chosen because they represent (i) active-probing formulations, (ii) robust risk-aware planning under multimodal uncertainty, and (iii) integrated probing and prediction approaches, i.e., they are the most relevant prior works to which our multimodal hierarchical, risk-aware probing framework can be fairly compared. To the best of our knowledge, none of these prior works combine a multimodal hierarchical intent–mode belief representation with an explicit tail-risk (CVaR) adjustment of multimodal weights and an entropy-minimizing probing objective; the three baselines above are therefore the closest existing methods.

Table~\ref{lane} (lane merge) and Table~\ref{4way} (unsignaled four-way intersection) summarize the quantitative comparison over 500 runs per planner. 
We observe our method outperforms consistently across all parameters. It is to be noted that the headway gap is less in our method, which is is deliberate and safe; it reflects an efficiency-driven decision. Our CVaR-based evaluation explicitly penalizes high-tail-cost outcomes, so the controller reduces conservative spacing when the tail risk is acceptable, producing faster merges while keeping worst-case outcomes constrained.
%For statistical validation we performed two-proportion $z$-tests for success rates and two-sided Welch $t$-tests for continuous metrics (all tests use $n=500$ per planner). 
%Key findings are:
The baselines differ in performance based on how they address uncertainty, probing, and risk. AP-IH focuses on information gain through probing but ignores risk and intent inference, resulting in safe yet slow maneuvers with lower success rates under high uncertainty. CC-MPC improves by incorporating risk-aware planning with chance constraints and Gaussian mixtures, but its assumption of fixed uncertainty makes it overly cautious and less adaptable. AP-MP further reduces ambiguity by combining prediction and probing, yet its flat intent model and lack of tail-risk handling leave it vulnerable to rare but critical failures.
Our method surpasses these baselines by updating hierarchical beliefs, shaping interactions through probing and influence, and adjusting for tail risk with CVaR. This enables the ego vehicle to safely gather and act on information, achieving faster, more confident maneuvers without compromising safety. The results highlight the importance of coupling belief updates with action for effective navigation in uncertain interactive driving.

Fig. 3 illustrates qualitative trajectories for the same example trial lane merge scenario (with the same parameters) across all methods. 
%\yl{should mention this is how all methods behave in one example trial}
We observe that AP-IH (brown) adopts a highly conservative approach, requiring longer time to merge; CC-MPC (pink) manages to merge somewhat faster but exhibits indecision and prolonged lateral alignment with human vehicles; and AP-MP (blue) achieves smoother motion but still requires more time to complete the merge. In contrast, our method (yellow) executes a confident and decisive maneuver, merging more quickly and with smoother transitions. This visualization aligns with the quantitative findings in Table~\ref{lane}, highlighting how our probing-driven belief updates enable faster resolution of interactions, fewer deadlocks, and higher overall efficiency.
%\yl{move this paragraph to the end of this section} dense traffic with very high traffic speed, which makes it difficult for ego to take meaningful probe actions
While our approach achieves strong overall performance, we observe failures in certain edge cases such as dense traffic with very high traffic speed, which makes it difficult for the ego to take meaningful probe actions.
In these situations, our probing actions may not sufficiently disambiguate aggressive human intent, causing the ego to mispredict behaviors and commit to unsafe trajectories, sometimes resulting in crashes or incomplete maneuvers. Therefore, future work includes exploring multi-step lookahead probing strategies to improve robustness.

% To overcome this, we aim to enhance our framework with richer intent modeling and multi-step lookahead probing strategies, enabling more robust handling of adversarial or ambiguous interactions.
%\yl{I am commenting this paragraph out since this sounds more like a conclusion paragraph.} %The experiments show that combining (i) a multimodal hierarchical intent–mode belief representation, (ii) CVaR-based tail-risk adjustment of multimodal weights, and (iii) an entropy-minimizing active probing objective yields a planner that (a) handles uncertainty better than the closest active- and risk-aware baselines, (b) adapts to human-driven agents by shaping beliefs through informative probing, and (c) balances risk and efficiency in a principled manner. These improvements support our central claim: active, risk-aware information-seeking integrated tightly with planning produces safer, faster, and more interpretable interactive maneuvers than prior approaches that treat prediction and planning as largely separate modules.
% \vspace{-.7cm}
\section{Conclusions}
\label{sec:conclusions} 
% \vspace{-.5cm}

We propose a principled framework that establishes a bidirectional feedback loop between control and prediction in interactive driving. Specifically, we introduce a risk-aware multimodal covariance steering method that actively learns and shapes human behavior, modeled through a hierarchical belief framework for intent inference. By proactively reducing ambiguity in predicting human actions, our approach improves safety and efficiency in scenarios such as lane merging and unsignaled intersections.

{\small
	\bibliographystyle{ieeetr}
	\bibliography{root.bib}
}

\end{document}